\documentclass[sigconf,nonacm]{acmart}

\usepackage{subcaption}
\usepackage{xspace}
\usepackage{listings}
\usepackage{enumitem}
\usepackage{multirow}
\usepackage[group-separator={,}]{siunitx}
\usepackage{makecell}

\AtBeginDocument{%
  }

\renewcommand\footnotetextcopyrightpermission[1]{}

\begin{document}

\setlength{\abovecaptionskip}{5pt plus 1pt minus 2pt}
\setlength{\textfloatsep}{5pt plus 1.0pt minus 2pt}
\setlength{\dbltextfloatsep}{5pt plus 1pt minus 2pt}



\title{Cross-Source Reasoning-based Correction for Author Name Disambiguation}

\author{Fanjin Zhang}
\authornote{Equal contribution.}
\affiliation{%
  \department{School of Information \\Engineering Research Center of Database and Business Intelligence\\Key Laboratory of Data Engineering and Knowledge Engineering}
  \institution{Renmin University of China }
  \city{Beijing}
  \country{China}}
\email{fanjinz@ruc.edu.cn}

\author{Yunhe Pang}
\authornotemark[1]
\authornote{Work was done when Yunhe interned at Z.ai.}
\affiliation{%
  \department{School of Computer Science and Engineering}
  \institution{Sun Yat-Sen University}
  \city{Guangzhou}
  \country{China}}
\email{pangyh8@mail2.sysu.edu.cn}

\author{Bo Chen}
\affiliation{%
\department{Department of Computer Science and Technology}
  \institution{Tsinghua University}
  \city{Beijing}
  \country{China}}
\email{allanchen224@gmail.com}

\author{Zhiyu Shen}
\affiliation{%
\department{School of Computer Science and Engineering}
  \institution{Sun Yat-Sen University}
  \city{Guangzhou}
  \country{China}}
\email{shenzhy23@mail2.sysu.edu.cn}

\author{Yanghui Rao}
\authornote{Yanghui Rao and Jie Tang are the corresponding authors.}
\affiliation{%
\department{School of Computer Science and Engineering}
  \institution{Sun Yat-Sen University}
  \city{Guangzhou}
  \country{China}}
\email{raoyangh@mail.sysu.edu.cn}

\author{Evgeny Kharlamov}
\affiliation{%
\institution{Robert Bosch GmbH}
\department{Bosch Center for AI}
\city{Renningen}
\country{Germany}
  }
\affiliation{
\institution{University of Oslo}
\department{Department of Informatics}
\city{Oslo}
\country{Norway}
}
\email{Evgeny.Kharlamov@de.bosch.com}

\author{Jie Tang}
\authornotemark[3]
\affiliation{%
\department{School of Computer Science and Engineering}
  \institution{Tsinghua University}
  \city{Beijing}
  \country{China}}
\email{jietang@tsinghua.edu.cn}
%

\lstdefinestyle{prompt}{
    basicstyle=\ttfamily\small,
    backgroundcolor=\color{gray!10},
    frame=single,
    breaklines=true,
    columns=fullflexible,
    keepspaces=true,
    showstringspaces=false
}
\renewcommand{\shortauthors}{Fanjin Zhang et al.}

\newcommand{\hide}[1]{}
\newcommand{\vpara}[1]{\vspace{0.07in}\noindent\textbf{#1 }}

\newcommand{\zfj}[1]{\textbf{\color{orange}[(ZFJ: #1 )]}}  

\newcommand{\pyh}[1]{\textbf{\color{blue}[(PYH: #1 )]}}  
\newcommand{\todo}[1]{\textbf{\color{red}[(TODO: #1 )]}}  
\newcommand{\model}{CrossND\xspace}

\newcommand{\cons}{consistent\xspace}
\newcommand{\Cons}{Consistent\xspace}
\newcommand{\incons}{inconsistent\xspace}
\newcommand{\Incons}{Inconsistent\xspace}
\newcommand{\fuzzy}{fuzzy\xspace}
\newcommand{\Fuzyy}{Fuzzy\xspace}

\newcommand{\mcs}{cs}
\newcommand{\mics}{ics}
\newcommand{\mfz}{fz}

\definecolor{comment}{RGB}{70, 150, 60}

\newcommand{\ain}{a^{\text{in}}}
\newcommand{\aout}{a^{\text{out}}}

\newcommand{\beq}[1]{\vspace{-0.1in}\begin{equation}#1\end{equation}\vspace{-0.1in}}

\begin{abstract}
Author name disambiguation is a critical challenge in academic search systems, often addressed through from-scratch and real-time disambiguation approaches. However, current algorithms remain vulnerable to cumulative errors of paper-author assignments and overlook inconsistent assignments across different sources. Resorting to expert annotation is resource-intensive. To this end, this paper explores a new perspective for author name disambiguation: cross-source correction by leveraging inconsistent assignments across sources. We propose CrossND, a full-stack framework that integrates data refinement, cross‑source reasoning, and test-time scaling. First, a chain‑of‑refinement pipeline denoises author profiles and produces more accurate paper-author matching probabilities. Second, a supervised fine‑tuning process incorporates these refined signals and a probabilistic soft logic-based cross‑correction module to infer the assignments of which sources are incorrect. Third, test-time scaling further enhances the accuracy and robustness of the predictions. Experiments on real‑world datasets indicate that CrossND consistently outperforms $17$ baselines by leveraging cross-source reasoning without human intervention. CrossND has been deployed to support large‑scale paper-author assignment correction in practice. 
\end{abstract}




\maketitle

\newcommand\codeavailabilityurl{https://github.com/zfjsail/CrossND}
\ifdefempty{\codeavailabilityurl}{}{
\begingroup\small\noindent\raggedright\textbf{Code Availability:}\\
The source code of this paper has been made publicly available at \url{\codeavailabilityurl}.
\endgroup
}

\section{Introduction}

Name disambiguation, which aims to distinguish among different individuals sharing the same name, is a fundamental component of academic search platforms~\cite{zhang2024oag},
such as Google Scholar, Web of Science, etc. Recently, the AI-in-the-loop paradigm has accelerated the pace of academic publishing significantly but brought great challenges to author name disambiguation as the inaccurate disambiguation results may lead to negative social impacts, e.g., invalid author rankings~\cite{schulz2016using} and award cheating~\cite{chen2022gccad}.

\begin{figure}[t]
	\centering
	\includegraphics[width=8cm]{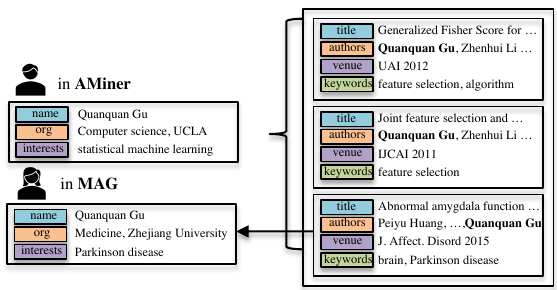}	
	\caption{A real motivating example. 
		In AMiner, Quanquan Gu at UCLA authored three papers on the right.
		However, the bottom paper is associated with Quanquan Gu at Zhejiang University in MAG.
		On closer examination, the top two and bottom papers belong to significantly different fields 
		and thus belong to different individuals named Quanquan Gu.
	}
	\label{fig:na-example}
\end{figure}

Existing endeavors mainly focus on From-Scratch Name Disambiguation,
which aims to partition papers associated with the same name into distinct clusters.
However, the imperfectness of name disambiguation algorithms results in inevitable cumulative errors. 
Moreover, inconsistent paper-author assignments are also prevalent across different systems. According to the statistics of the WhoIsWho~\cite{chen2023web}\footnote{\url{https://www.aminer.cn/open/article?id=5de9efd2530c707ed8b87d99}} dataset, 
\textbf{the ratio of inconsistent assignments is about} $\mathbf{10\%}$ between AMiner~\cite{tang2008arnetminer} and MAG~\cite{sinha2015overview}.
Figure \ref{fig:na-example}, a real motivating example, demonstrates the difference of the paper ownership by Quanquan Gu in AMiner and MAG,
implying that more credible decisions on paper assignments may be made by employing external knowledge.

Recent attempts commonly rectify name disambiguation results in a single source. Some works transform this problem into a graph anomaly detection task by constructing paper similarity graphs. Take GCCAD~\cite{chen2022gccad} as an example, it compares each node to the global context to produce more discriminative node embeddings.
More recent studies~\cite{pang2025guard,zhang2024enhanced} propose tailored instruction formats and fine‑tune large language models (LLMs) for anomaly detection, yielding better results than conventional approaches.
However, these methods heavily rely on human-annotated incorrect paper-author assignments as anomalies for training. Moreover, these methods overlook the fact that naturally occurring external paper-author assignments can serve as supplementary information for reasoning and error correction.

To this end, we investigate a new name disambiguation paradigm,
i.e., the cross-source correction.
Considering the powerful reasoning capabilities of LLMs, 
we explore their ability to perform cross-source reasoning and rectify errors.
Intuitively, we can first measure paper-author matching similarity, and then identify inconsistent paper-author matchings across sources to detect the incorrect ones.
Such a strategy poses the following challenges. Firstly, LLMs are prone to overfitting noisy labels due to their high parameter count and strong expressive power.
Secondly,
deciding the paper-author assignments of \textbf{which sources are incorrect}
is difficult when dealing with inconsistent assignments.

Accordingly, we propose the \model framework to tackle the aforementioned challenges: 
1) A chain‑of‑refinement pipeline is performed to clean author profiles and leverage external data to generate refined author-paper matching probabilities, thereby mitigating the impact of noisy labels.
2) We develop a  supervised fine-tuning (SFT) framework that absorbs refined labels and a cross-correction method based on probabilistic soft logic
to infer paper-author pairs of which sources are incorrect.
3) A simple bootstrapping test-time scaling strategy that integrates cross-source data clearly improves prediction accuracy and robustness.
Our contributions include:

\begin{itemize}[leftmargin=*]
	\item We present a new perspective for name disambiguation --- 
	cross-source correction,
	which is aware of noisy labels 
	and can improve name disambiguation without human intervention 
	by incorporating external knowledge.
	\item We design a full-stack framework \model to address the challenges stated above. We first develop a chain-of-refinement data distillation pipeline to combat noisy labels. Afterward, we propose a cross-correction method to infer the incorrect paper-author pairs from inconsistent assignments and inject it into the multi-turn SFT objective.

	\item Our experiments show that \model consistently outperforms various baselines, 
	manifesting the effectiveness of 
	our design choice for each component in our framework.
	\model has been deployed to assist human verification for name disambiguation by leveraging multiple data sources\footnote{\url{https://na-demo.aminer.cn/crosscheckauthor}}.

\end{itemize}
\section{Problem Definition}

Given a set $G^{\text{in}} = \{P,  A^{\text{in}}, R^{\text{in}}\}$ in the internal source
and $G^{\text{out}} = \{P, A^{\text{out}}, R^{\text{out}}\}$ in an external source,
where $P$ is the paper set, 
$A^{\text{in}}$/$A^{\text{out}}$ is the author set in the internal/external source,
and $R^{\text{in}}$/$R^{\text{out}}$ represents the set of authorship relations 
between authors and papers in $G^{\text{in}}$/$G^{\text{out}}$,
we propose to study to what extent existing paper-author assignments 
can be rectified by comparing and cross-checking two sources. 
Note that we assume paper entities are well-disambiguated between the two sources, 
as most papers are identifiable through their titles and author lists.

This work aims to discover potential incorrect paper-author pairs.
The input dataset is defined as $\mathcal{D} = \{d = (p, a^{\text{in}}, a^{\text{out}}, \kappa) | p \in P, a^{\text{in}} \in A^{\text{in}}, $ $a^{\text{out}} \in A^{\text{out}}, \kappa \in \mathbb{N}^+)\}$,
where $p$ is the target paper 
from the paper set $P$,
$a^{\text{in}}/a^{\text{out}}$ is the $\kappa$-th author of paper $p$ 
in the author set $A^{\text{in}}/A^{\text{out}}$,
and $\kappa$ denotes the author order in paper $p$, 
which is omitted in the following if there is no ambiguity.
Each paper has multiple attributes, such as title, venue, and keywords.
By representing each author as a list of papers, 
author $a^{\text{in}}$ is denoted as $a^{\text{in}} = [p_1, p_2, ..., p_n]$,
with $n$ being the count of papers authored by $a^{\text{in}}$.
We view each author as a combination of one's papers 
since other attributes, such as keywords and co-authors, are derived from the papers.
We aim to learn a matching function $\mathcal{M}(p, a^{\text{in}} , a^{\text{out}}) \to [0, 1]$ 
to decide the correctness of the paper-author pair $(p, a^{\text{in}})$. 
\section{Model Framework}

\begin{figure*}[t]
	\centering
	\includegraphics[width=0.95\textwidth]{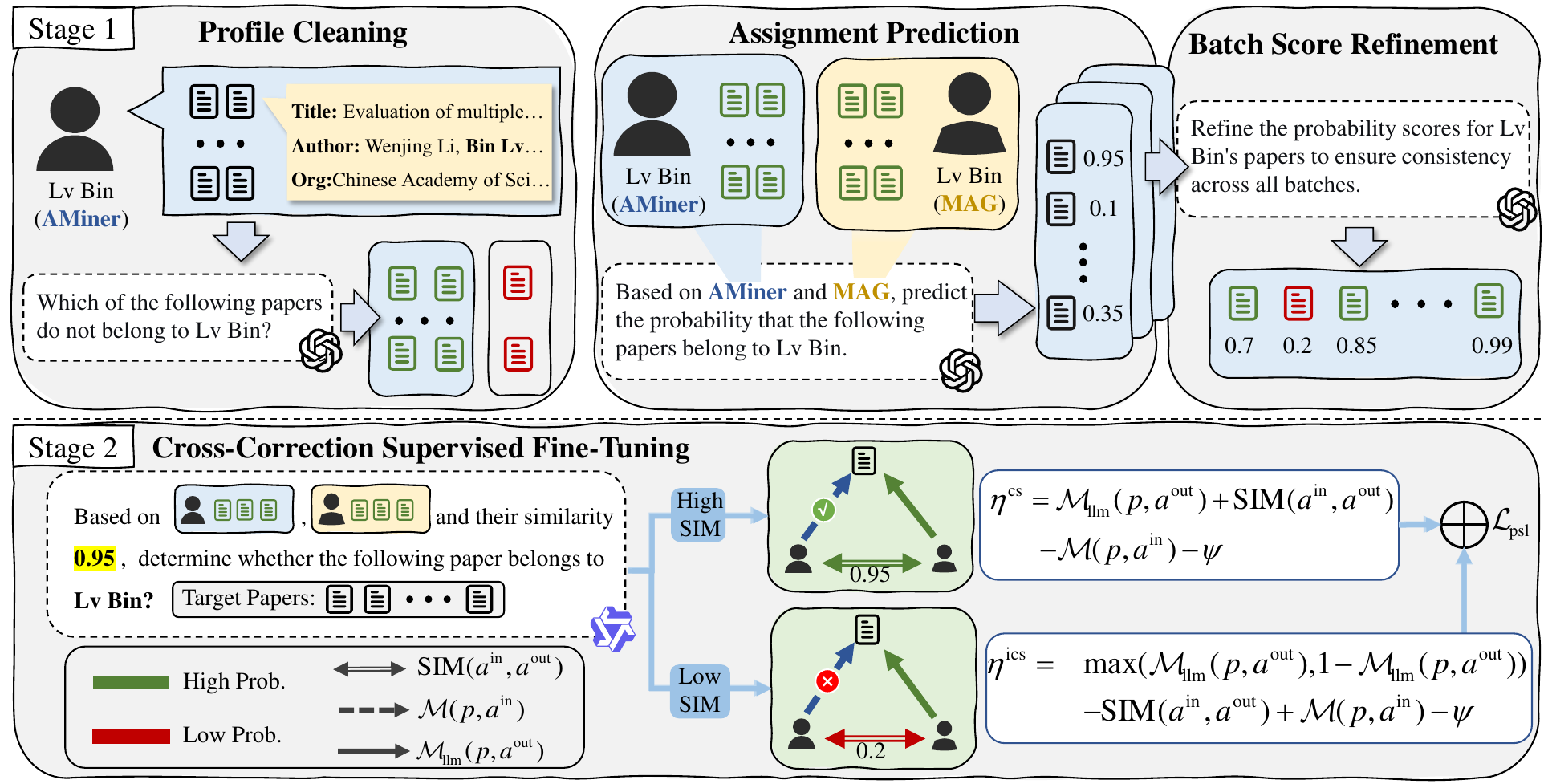}	
	\caption{Overall architecture of CrossND. \normalfont  The upper part shows the Chain-of-Refinement Pipeline, which distills high-quality soft labels from an LLM through three stages: profile cleaning, assignment prediction, and batch score refinement. The lower part presents the Cross-Correction Supervised Fine-Tuning, which partitions training data via cross-source author similarity and incorporates a PSL-based loss to impose structured logical constraints among paper-author triples.}
	\label{fig:model}
\end{figure*}

Previous studies on author name disambiguation primarily rely on a single data source, which limits their ability to identify inconsistent author–paper assignments in complex academic systems.
To address the limitations of single-source methods and the challenge of inherent label noise, we propose \model, a full-stack pipeline that integrates label denoising, cross-correction fine-tuning, and test-time scaling for cross-source correction in author name disambiguation.
Firstly, our framework begins with a carefully designed chain-of-refinement pipeline to mitigate label noise, including profile cleaning, assignment prediction, and batch score refinement.
Secondly, we introduce a cross-source supervised fine-tuning model and a cross-correction method based on probabilistic soft logic to deduce the incorrect papers through the modified objective.
Thirdly, we employ a bootstrapping-based test-time scaling mechanism that integrates information from multiple inference passes, yielding more accurate and robust predictions.
The overall framework is illustrated in Figure~\ref{fig:model}.





\subsection{Chain-of-Refinement Pipeline}

We propose a three-step pipeline to mitigate the label noise of paper-author assignments, including profile cleaning, assignment prediction, and batch score refinement.
(1) In the profile cleaning step, we extract a representative coreset for each author in order to remove potential incorrect paper-author assignments.
(2) In the assignment prediction step, we obtain soft prediction scores from a decent LLM API (e.g. GPT-5) based on the target papers and the cleaned author profiles.
(3) In the batch score refinement step, we further make batch predictions from the same LLM API by providing preceding prediction scores to address inconsistency scores across batches of API calls.

\vpara{Profile Cleaning.}
This step aims to identify a coreset from the original author profile, which may contain incorrect paper assignments. The goal of this step is to retain only the papers that truly reflect the author's research identity.

Formally, given an author profile $a = \{p_1, p_2, ..., p_n\}$ containing $n$ papers, we employ an off-the-shelf LLM to compute a binary anomaly decision for each paper in the coreset:
\begin{equation} 
\overset{*}{y}_i = \text{LLM}_{\text{clean}}(p_i, a), \quad \overset{*}{y}_i \in \{0,1\}, \; i \in \{1,2,...,n\} 
\end{equation}

where $\overset{*}{y}_i = 1$ means that paper $p_i$ is judged to be consistent with the author's profile, and $\overset{*}{y}_i = 0$ otherwise. We then construct the coreset by selecting only the positive papers:  
\begin{equation} a^{\text{core}} = \{p_i \in a \mid \overset{*}{y}_i = 1\}. 
\end{equation}



The LLM first models an author’s publications to capture dominant research topics, then evaluates individual papers in context to identify deviations. This contextual reasoning enables more accurate discrimination between genuine and misattributed papers.


This denoising step is essential for downstream tasks. By removing anomalous papers early in the pipeline, we obtain a cleaner and more faithful coreset that captures the author's true expertise, thereby reducing noise and improving the reliability of subsequent label prediction and knowledge distillation stages.


\vpara{Assignment Prediction.}
\label{sec:assign_pred}
To generate the refined training labels, we employ the same API-based LLM to predict soft assignment scores using information from two sources: the internal author $a^{\text{in}}$
and the external author $a^{\text{out}}$. For each paper $p_i$, the LLM is prompted to produce a soft score.

However, constructing long LLM contexts for each paper-author pair incurs significant computational and API overhead. 
To address this, we adopt a batch querying strategy~\cite{cheng2023batch}. Papers are grouped into batches $\mathcal{B} = \{p_{1}, ..., p_{m}\}$, and the LLM jointly predicts assignment scores for all papers in the batch:
\begin{equation}
\{\hat{y}_{1}, ..., \hat{y}_{m}\} = \text{LLM}_{\text{pred}}(\mathcal{B}, a^{\text{in}}, a^{\text{out}})
\end{equation}

Batching significantly reduces the number of API calls while preserving prediction quality, enabling the LLM to produce reliable soft labels that serve as valuable signals for subsequent refinement.





\vpara{Batch Score Refinement.}
However, we observe that LLM inference can produce inconsistent soft-label ranges across batches, even when outputs are restricted to the interval $[0, 1]$. As a result, scores for papers from the same author may vary noticeably across batches, introducing undesirable calibration noise. To address this issue, by employing the same LLM, we aggregate all preliminary predictions for an author $a^{\text{in}}$, denoted by $\{\hat{y}_1, ..., \hat{y}_n\}$, and apply a refinement step that enforces global consistency:
\begin{equation}
\{\tilde{y}_1, ..., \tilde{y}_n \}= \text{LLM}_{\text{refine}}(\{(p_1, \hat{y}_1), ..., (p_n, \hat{y}_n)\})
\end{equation}

\noindent where $p_i$ is the paper associated with score $\hat{y}_i$, and $\tilde{y}_i$ denotes the calibrated scores.

This refinement step leverages the global score distribution to correct batch-wise variance while preserving the relative ordering of predictions. The resulting harmonized soft labels exhibit improved cross-batch consistency and serve as the final distilled supervision signals for downstream fine-tuning.
Detailed prompts in this pipeline are provided in Appendix \ref{app:appendix_prompt}.



\hide{
The refinement process takes into account the global distribution of scores to produce calibrated labels:
\begin{equation}
\mathbf{y}^{\text{refined}} = f_{\text{refine}}(\hat{\mathbf{y}}, \{p_1, ..., p_n\})
\end{equation}
where $\hat{\mathbf{y}} = [\hat{y}_1, ..., \hat{y}_n]$ represents all preliminary scores and $\mathbf{y}^{\text{refined}}$ denotes the refined, consistency-calibrated scores.

This refinement process effectively mitigates the batch-wise variance inherent in LLM predictions and produces a harmonized set of soft labels that maintain relative ordering while ensuring cross-batch calibration. The refined scores serve as the final distilled labels for subsequent supervised fine-tuning.
}

\subsection{Cross-Correction Supervised Fine-Tuning}

From the previous steps, we obtain author-paper matching scores calibrated by the LLMs. Determining whether a paper is incorrectly attributed to an author can be formulated as an anomaly detection task. Prior work has demonstrated that fine‑tuning LLMs can yield strong anomaly detection performance. Among these methods, GuARD~\cite{pang2025guard} employs a pseudo multi‑turn fine‑tuning strategy that leverages shared context to jointly predict multiple samples, achieving both high accuracy and computational efficiency. We adopt a similar idea and adopt a multi-turn SFT method.

A straightforward method is to minimize the cross‑entropy loss. However, 
this ignores the logical dependencies 
that exist among the triple $(p, a^{\text{in}} , a^{\text{out}})$. In fact, the decision of whether two entities match is conditioned on the third entity in the triple. To resolve this, we incorporate a probabilistic soft logic (PSL)–based framework~\cite{kimmig2012short} that explicitly models these inter‑pair dependencies. Our method first groups data according to cross-source author similarity, and then formulates a PSL‑driven loss function that captures the structured matching constraints embedded in the triples.

\vpara{Consistent-inconsistent Data Partitioning.}
We consider triples $(p, a^{\text{in}}, a^{\text{out}})$ 
and measure the similarity between $a^{\text{in}}$ and $a^{\text{out}}$, denoted as $\text{SIM}(a^{\text{in}}, a^{\text{out}})$.
A high $\text{SIM}(a^{\text{in}}, a^{\text{out}})$ implies that 
the relation $(p, a^{\text{in}})$ is acknowledged by external knowledge,
while a low one suggests that \textbf{at least one of} 
$(p, a^{\text{in}})$ and $(p, a^{\text{out}})$ is incorrect.

$\text{SIM}(a^{\text{in}}, a^{\text{out}})$ can be estimated in various ways~\cite{liu2016aligning,zhang2019oag}. To improve the prediction accuracy, we adopt a simple yet robust statistical cue to distinguish similar from dissimilar authors. 
Since publications are the most reliable author attribute, we primarily rely on paper-level evidence rather than derived attributes such as co-venues or co-authors. 
Specifically, we define the similarity between two authors based on their publication overlap as
\begin{equation}
\label{eq:paper_overlap}
\text{SIM}(a^{\text{in}}, a^{\text{out}})=
\min\!\left(
\frac{|P(a^{\text{in}})\cap P(a^{\text{out}})|}{|P(a^{\text{in}})|},
\frac{|P(a^{\text{in}})\cap P(a^{\text{out}})|}{|P(a^{\text{out}})|}
\right),
\end{equation}

\noindent where $P(a)$ denotes the set of papers authored by $a$. 
According to author similarities, 
the training set $\mathcal{D}$ can be partitioned into two subsets:
\cons set $\mathcal{D}_{\text{\mcs}}$ and \incons set $\mathcal{D}_{\text{\mics}}$,
which are defined as follows:

\beq{\nonumber
	\left\{
	\begin{aligned}
		&\mathcal{D}_{\text{\mcs}} =  \{(p, a^{\text{in}}, a^{\text{out}})| \text{SIM}(a^{\text{in}}, a^{\text{out}}) > \rho\}\\
		&\mathcal{D}_{\text{\mics}} =  \{(p, a^{\text{in}}, a^{\text{out}})| \text{SIM}(a^{\text{in}}, a^{\text{out}}) \leq \rho\}
	\end{aligned}
	\right.
}

\vpara{Probabilistic Soft Logic (PSL).}
We estimate pairwise matching probabilities within each triple using LLMs or other models. We aim to use the matching probabilities of $(a^{\text{in}}, a^{\text{out}})$ and $(p, a^{\text{out}})$ to infer the matching probability of $(p, a^{\text{in}})$.
To capture such dependencies, 
we make use of Probabilistic Soft Logic (PSL) \cite{kimmig2012short}, a compact reasoning framework that supports soft logical relationships.

PSL represents each fact or rule with a continuous truth value in [0,1], reflecting their degrees of belief.
Logical operators (such as AND, OR, NOT) are relaxed into differentiable functions using the Lukasiewicz t‑norm, which enables soft reasoning instead of hard Boolean decisions. For example, the truth value of a conjunction is computed as $u_1 \land u_2 = \max\{0, \mathcal{I}(u_1) + \mathcal{I}(u_2) - 1 \}$, where $\mathcal{I}(u)$ is the confidence score of a rule $u$. 

Each rule in PSL, such as
$\gamma$: $u_1 \land u_2 \to u_3$,
is considered satisfied when the confidence of the head ($u_3$) is at least as large as the confidence implied by its body ($u_1 \land u_2$). Thus, a difference value $\eta_\gamma$ can be defined as:

\beq{
\eta_\gamma  =  \mathcal{I}(u_1 \land u_2) - \mathcal{I}(u_3) \\
= \mathcal{I}(u_1) + \mathcal{I}(u_2) - 1 - \mathcal{I}(u_3)
}

\noindent where $\gamma$ is satisfied\footnote{Here we leave out the trivial case of $\mathcal{I}(u_1 \land u_2) = 0$.} when the difference $\eta_\gamma \leq 0.$

Overall, PSL provides a principled way to propagate confidence from observed pairwise matchings to the unobserved triple‑level relation. By enforcing these soft logical dependencies, the model becomes more consistent and captures the mutual constraints among the three pairs in each triple.

\vpara{Cross-correction via PSL.}
To be specific, we aim to deduce the matching of $(p, \ain)$ from the matchings of $(p, \aout)$ and $(\ain, \aout)$.
For instances in \cons set $\mathcal{D}_{\text{\mcs}}$, 
we define $v_1 = \mathcal{M}_{\text{llm}}(p, \aout)$, 
$v_2 = \text{SIM}(\ain, \aout)$, and $\mathcal{I}(u_3) = \mathcal{M}(p, \ain)$,
where $\mathcal{M}_{\text{llm}} (\cdot)$ uses the off-the-shelf LLM to obtain matching probability like Section ~\ref{sec:assign_pred}, and $\mathcal{M} (\cdot)$ denotes the output logits of the finetuned LLM.
Thus, for instances in $\mathcal{D}_{\text{\mcs}}$, we calculate:

\hide{
After... . In this section, we introduce a Probabilistic Soft Logits(PSL) framework for model optimization
\subsubsection{Probabilistic Soft Logits}
The aforementioned cross-source error correction method does not explicitly model the similarity between cross-source author pairs $(a^{\text{in}}, a^{\text{ext}})$. Here we consider an alternative perspective—treating the triplet $(p, a^{\text{in}}, a^{\text{out}})$ as a logical rule, since matching any two entities in the triplet depends on the remaining entity.

Given that we can now compute matching probabilities for entity pairs, such as $(a^{\text{in}}, a^{\text{out}})$ and $(p, a^{\text{out}})$, we introduce Probabilistic Soft Logic (PSL) to model the dependency among pairwise entity matches in the triplet. PSL is a probabilistic reasoning framework that propagates known belief probabilities to unknown rules. PSL assigns each rule a value between 0 and 1 indicating the probability that the rule holds true.

\begin{equation}
\mathcal{I}(u) = v_u \quad (0 \leqslant v_u \leqslant 1)
\end{equation}

where $\mathcal{I}(u)$ represents the probability that rule $u$ holds, with value $v_u$. In the PSL framework, the Łukasiewicz t-norm is used to define basic logical operations:

\begin{equation}
\begin{aligned}
u_1 \land u_2 &= \max\{0, \mathcal{I}(u_1) + \mathcal{I}(u_2) - 1\}, \\
u_1 \lor u_2 &= \min\{1, \mathcal{I}(u_1) + \mathcal{I}(u_2)\}, \\
\neg u_1 &= 1 - \mathcal{I}(u_1).
\end{aligned}
\end{equation}

PSL considers a rule to hold when the probability of the rule head is no lower than the probability of its body. For rule $\gamma: u_1 \land u_2 \rightarrow u_3$, we define the difference value $\eta_\gamma$ as:

\begin{equation}
\eta_\gamma = \mathcal{I}(u_1 \land u_2) - \mathcal{I}(u_3) = v_1 + v_2 - 1 - \mathcal{I}(u_3)
\end{equation}

where rule $\gamma$ holds when $\eta_\gamma \leqslant 0$.

Specifically, the PSL-based cross-source error correction method aims to infer the correctness of match $(p, a^{\text{in}})$ by leveraging the matching results of $(p, a^{\text{out}})$ and $(a^{\text{in}}, a^{\text{out}})$. For samples in the consistent set $\mathcal{D}_{\text{cs}}$, we define $v_1 = \mathcal{P}_{\text{old}}(p, a^{\text{out}})$, $v_2 = \text{SIM}(a^{\text{in}}, a^{\text{out}})$, and $\mathcal{I}(u_3) = \mathcal{P}(p, a^{\text{in}})$, where $\mathcal{P}_{\text{old}}(\cdot)$ and $\mathcal{P}(\cdot)$ have the same meaning as in the comparison-based cross-source error correction method. Thus, for samples in the consistent set, we compute:
}

\begin{equation}
\eta^{\text{cs}} = \mathcal{M}_{\text{llm}}(p, a^{\text{out}}) + \text{SIM}(a^{\text{in}}, a^{\text{out}}) - \mathcal{M}(p, a^{\text{in}}) - \varphi
\label{eq:psl_cs}
\end{equation}

\noindent where $\varphi$ is a flexible margin hyperparameter. In Eq. (\ref{eq:psl_cs}), $\mathcal{M}(\cdot)$ contains the parameters to be optimized during training.

In contrast, the inconsistent set $\mathcal{D}_{\text{ics}}$ presents a more complex scenario. For each triplet, at least one of the paper-author matches $(p, a^{\text{in}})$ and $(p, a^{\text{out}})$ is incorrect, or both are incorrect. To jointly model these possibilities, the difference value for the inconsistent set $\mathcal{D}_{\text{ics}}$ is defined as:

\begin{align}
\label{eq:psl-neg}
    \centering
    \begin{split}
        \eta^{\text{\mics}}  
        = &  \max(\mathcal{M}_{\text{llm}}(p, a^{\text{out}}), 1-\mathcal{M}_{\text{llm}}(p, a^{\text{out}}) )  \\
        & + (1 - \text{SIM}(a^{\text{in}}, a^{\text{out}}))
         - (1 - \mathcal{M}(p, a^{\text{in}})) - \varphi \\
        = & \max(\mathcal{M}_{\text{llm}}(p, a^{\text{out}}), 1 - \mathcal{M}_{\text{llm}}(p, a^{\text{out}}) ) 
        - \text{SIM}(a^{\text{in}}, a^{\text{out}}) \\
        & + \mathcal{M}(p, a^{\text{in}}) -\varphi 
    \end{split}
\end{align}

In Eq. (\ref{eq:psl-neg}), $\mathcal{M}(p, a^{\text{in}})$ denotes the soft-normalized score derived from the log-probabilities  ``Yes'' and ``No'' tokens as follows: 

\beq{
\mathcal{M}(p, a^{\text{in}}) = \frac{\exp(\text{logprob}_{\text{Yes}})}{\exp(\text{logprob}_{\text{Yes}}) + \exp(\text{logprob}_{\text{No}})}
}

\noindent where 
$1 - \mathcal{M}(p, a^{\text{in}})$ indicates the mismatched probability of pair $(p, a^{\text{in}})$.
Since we don't know whether pair $(p, a^{\text{out}})$ is matched or not, 
we adopt $\max(\mathcal{M}_{\text{llm}}(p, a^{\text{out}}), 1-\mathcal{M}_{\text{llm}}(p, a^{\text{out}}) )$ to cover the two cases.
In this way, 
large $\eta^{\text{\mics}}$ might imply correct internal pair $(p, a^{\text{in}})$ and incorrect external pair $(p, a^{\text{out}})$, 
while small $\eta^{\text{\mics}}$ possibly means incorrect internal pair $(p, a^{\text{in}})$.
Here, we also use the same margin $\varphi$ as that in $\eta^{\text{\mcs}}$.

The objective function for the PSL-based cross-source correction method combines a PSL loss with a cross-entropy loss:
\[
\begin{aligned}
&\mathcal{L}_{\text{psl}} =
\frac{1}{N_e} \sum_{i=0}^{N_e}
\Bigl(y_i \min(0, \eta_i^{\text{cs}})
+ (1 - y_i) \min(0, \eta_i^{\text{ics}})\Bigr), \\[3pt]
&\mathcal{L}_{\text{ce}} =
-\frac{1}{N_e} \sum_{i=0}^{N_e}
\Bigl(y_i \log \mathcal{M}(p_i, a_i^{\text{in}})
+ (1 - y_i) \log (1 - \mathcal{M}(p_i, a_i^{\text{in}}))\Bigr), \\[3pt]
&\mathcal{L}_{\text{total}} =
\lambda \mathcal{L}_{\text{psl}} + (1 - \lambda) \mathcal{L}_{\text{ce}} .
\end{aligned}
\]
\hide{
\begin{equation}
\mathcal{L}_{\text{psl}} = \frac{1}{N_e} \sum_{i=0}^{N_e} \left(y_i \min(0, \eta_i^{\text{cs}}) + (1 - y_i) \min(0, \eta_i^{\text{ics}})\right)
\end{equation}

\begin{equation}
\mathcal{L}_{\text{ce}} = -\frac{1}{N_e} \sum_{i=0}^{N_e} \left(y_i \log \mathcal{M}(p_i, a_i^{\text{in}}) + (1 - y_i) \log (1 - \mathcal{M}(p_i, a_i^{\text{in}}))\right)
\end{equation}

\begin{equation}
\mathcal{L}_{\text{total}} = \lambda \mathcal{L}_{\text{psl}} + (1 - \lambda) \mathcal{L}_{\text{ce}}
\end{equation}
}
\noindent where $N_e = |\mathcal{D}_{\text{cs}} \cup \mathcal{D}_{\text{ics}}|$, $y_i = 1$ indicates the corresponding sample is in the consistent set, and $y_i = 0$ indicates it is in the inconsistent set. 
The hyperparameter $\lambda \in [0, 1]$ balances the contribution of the PSL-based loss and the cross-entropy loss.
The minimization operation $\min$ in $\mathcal{L}_{\text{psl}}$ is utilized to optimize samples with small $\eta$ values, as the corresponding rules for these samples are more likely to hold. Conversely, when $\eta^{\text{\mics}}$ or $\eta^{\text{cs}}$ are greater than zero, the rule is violated, suggesting that the sample may be noisy.

\hide{
\subsection{Test Time Scaling}
Despite the chain-of-refinement pipeline and cross-correction fine-tuning, we observe that inference still suffers from uncertainty arising from profile selection and multi-turn batch prediction. In particular, the coreset construction of author profiles and the ordering of papers in batch prediction can introduce non-negligible variance in the final soft scores, especially for borderline cases where the matching probability is close to the decision threshold.

To enhance prediction robustness, we further introduce a simple Test-Time Scaling (TTS) strategy. During inference, we perform multiple stochastic runs by randomly sampling subsets of the author profile and reshuffling the target papers before multi-turn prediction. The model’s final decision is obtained by aggregating predictions across multiple runs (e.g., averaging). This strategy effectively reduces sensitivity to profile composition and batch ordering, leading to more stable and reliable cross-source correction results.
}

\subsection{Test-Time Scaling}
\label{sec:tts}
Prior work~\cite{pang2025guard} predicts multiple papers via a multi-turn template but orders target papers randomly, limiting in-context calibration. We propose a Test-Time Scaling (TTS) strategy that reorders inputs by prediction confidence to provide stable contextual anchors.


\vpara{Confidence Estimation.}
We obtain soft predictions $\bar{y}_i \in [0,1]$ for each paper $p_i$. 
We define the confidence as the distance to the decision boundary:
$
c_i = \lvert \bar{y}_i - 0.5 \rvert .
$
Let $\mathcal{P}=\{p_{(1)},\dots,p_{(N)}\}$ denote the papers sorted by decreasing confidence, i.e., 
$c_{(1)} \ge \cdots \ge c_{(N)}$.

\hide{
\vpara{Confidence Estimation.} 
We first perform a preliminary pass to obtain initial soft scores $\hat{y}_i^{(1)} \in [0, 1]$ for all $N$ target papers associated with an author. We define the prediction confidence $c_i$ for each paper $p_i$ as the distance from the decision boundary:
\begin{equation}
c_i = |\hat{y}_i^{(1)} - 0.5|, \quad i \in \{1, \dots, N\}
\end{equation}
Let $\mathcal{P} = \{p_{(1)}, p_{(2)}, \dots, p_{(N)}\}$ be the set of papers sorted in descending order of confidence, such that their corresponding confidences satisfy $c_{(1)} \geq c_{(2)} \geq \dots \geq c_{(N)}$.
}

\vpara{Cross-batch balancing (snake dealing)}.
Papers $\mathcal{P}$ are then assigned to batches using a snake (zigzag) scheme: in the first round, papers are distributed from batch 1 to $B$, and in the next round from $2B$ back to $B+1$, etc. This ensures that each batch receives a similar mix of high- and low-confidence papers.

\vpara{Within-batch interleaving}.
For each batch, papers are sorted by descending confidence and then reordered in an alternating high-low pattern (highest, lowest, second-highest, second-lowest, …). This increases contextual confidence and diversity. By bootstrapping on high-confidence samples, the model leverages them as few-shot anchors to improve calibration on low-confidence cases.


\hide{
\begin{enumerate}
    \item \textbf{Global Serpentine Partitioning:} To ensure an even distribution of difficulty across batches, we assign sorted papers to batches using a serpentine mapping $\sigma(j)$. For the $j$-th paper $p_{(j)}$, the batch index $k \in \{1, \dots, K\}$ is determined by:
    \begin{equation}
    k = \begin{cases} 
    (j-1) \pmod K + 1 & \text{if } \lceil j/K \rceil \text{ is odd} \\
    K - (j-1) \pmod K & \text{if } \lceil j/K \rceil \text{ is even}
    \end{cases}
    \end{equation}
    This prevents any single batch from being dominated by low-confidence samples, which could otherwise lead to cumulative reasoning errors.

    \item \textbf{Intra-batch Interleaving:} Within each batch $B_k$, we reorder papers to alternate between high and low confidence. Let $\{q_1, q_2, \dots, q_M\}$ be the subset of papers assigned to $B_k$, re-indexed by descending confidence. The final input sequence $\mathcal{S}_k$ is constructed as:
    \begin{equation}
    \mathcal{S}_k = \{q_1, q_M, q_2, q_{M-1}, q_3, \dots\}
    \end{equation}
\end{enumerate}
}

\section{Experiments}
\subsection{Experimental Setup}
\vpara{Datasets.}
We adopt two widely-used publicly available name disambiguation datasets for incorrect author-paper assignment detection.
Table~\ref{tab:dataset_stats} provides an overview of the dataset statistics.

\begin{itemize}[leftmargin=*]
    \item \textbf{WhoIsWho~\cite{chen2023web}}: It
    is compared with the OAG snapshot~\cite{zhang2019oag,zhang2022oag} to identify historical AMiner assignment errors. For training, we upsample anomaly instances and obtain 381,772 author-paper pairs from the  AMiner
    to achieve a 1:1 positive-negative ratio.
    Ground‑truth instances are split by authors into validation and test sets. We also incorporate the MAG as external knowledge.
    \item \textbf{KDD Cup~\cite{roy2013microsoft}}: It
    includes user edits that flag incorrect author-paper assignments, which we treat as ground truth. We upsample anomaly instances and sample 31,997 author-paper pairs for training
    to achieve a 1:1 positive-negative ratio.
    Validation and test sets are split by authors. We use the AMiner release~\cite{zhang2019oag} as external knowledge. The WhoIsWho and KDD Cup datasets have minimal overlap in both authors and papers.
\end{itemize}

\begin{table}[t]
\centering
\caption{Statistics of the name disambiguation datasets.}
\label{tab:dataset_stats}
\begin{tabular}{lcc}
\toprule
\textbf{Statistics} & \textbf{WhoIsWho} & \textbf{KDD Cup} \\
\midrule
\# Training samples & \num{241740} & \num{31997} \\
\# Validation samples & \num{31873} & 4848 \\
\# Test samples     & \num{17136} & 5443 \\
\midrule
\# Val pos./neg. samples & \num{27539}/4334 & 4209/639 \\
\# Test pos./neg. samples & \num{14532}/2604 & 4663/780 \\
\% Anomaly & $14.16\%$ & $13.79\%$ \\
\bottomrule
\end{tabular}
\end{table}

\vpara{Evaluation Metrics.} 
Given the highly skewed distribution of positive and negative pairs, and following standard practice in anomaly detection, we report AUC as the primary evaluation metric. In addition, we use mean average precision (MAP), which emphasizes the ranking quality of incorrectly predicted instances. For each metric, we compute a macro average over all authors, as the difficulty of cross-correction varies substantially across different authors.

\vpara{Baselines.}
We compare various baselines:

\begin{itemize}[leftmargin=*]
    \item GBDT~\cite{friedman2001greedy} is a widely used ensemble model. Following CONNA~\cite{chen2022conna}, we use the same input features, including statistical signals extracted from titles, co-authors, affiliations, venues, and keywords.
    \item RND-all~\cite{chen2023web} follows the WhoIsWho contest winning solution, combining pre‑trained embeddings~\cite{liu2021oag} with key hand‑crafted similarity features, and applying multiple ensemble classifiers for prediction.
    \item AuthorSim uses cross-source author similarities as the match probability of paper-author pairs.
    \item LOF~\cite{breunig2000lof} is a feature‑based anomaly detector that measures each instance’s local deviation from its neighbors. We represent each paper using the average of its token embeddings.
    \item DCI~\cite{wang2021decoupling} is a graph‑based anomaly detector that builds a paper similarity graph for each author, learns node embeddings via graph partitioning, and computes anomaly scores from the similarity between node and cluster embeddings.
    \item KEnS~\cite{chen2020multilingual} embeds entities from multiple knowledge graphs into a shared space and uses ensemble inference across graphs to verify relation existence.
    \item TK~\cite{hofstatter2020interpretable} is a neural re-ranking model that uses Transformer layers for text contextualization and kernel aggregation for matching.
    \item CONNA~\cite{chen2022conna} is a state‑of‑the‑art on‑the‑fly name‑disambiguation method that uses a kernel‑based aggregation model to capture interactions across multiple fields and multiple instances.
\end{itemize}

The self‑correction baselines adopt CONNA as the backbone and employ the following modified loss functions.

\begin{itemize}[leftmargin=*]
    \item LS: utilizes the label smoothing loss as the objective function.
    \item MAE~\cite{ghosh2017robust}: uses an MAE-based loss, which is more robust to label noise than cross‑entropy.
    \item Co-teaching~\cite{han2018co}: trains two networks jointly, each guiding the other using selectively confident peer predictions.
    \item CORES~\cite{cheng2021learning}: leverages a novel confidence regularization term and progressively sieves out corrupted examples.
\end{itemize}

For LLM-based methods, we use DeepSeek‑R1‑0528,  Claude‑Haiku-4.5, GPT‑5, and Gemini-3-Pro-Preview as flagship models~\cite{guo2025deepseek,openai2025gpt5,anthropic2025claude45haiku}, applying only the first step of the chain‑of‑refinement pipeline before making direct predictions. Results for additional flagship models are reported in the Figure ~\ref{fig:llms_cmp}. We also compare against the GuARD framework~\cite{pang2025guard}, which performs pseudo-multi‑turn LLM fine‑tuning for anomaly detection tasks with text‑rich inputs. To verify the generality of PSL loss, we augment the objective function of the CONNA method by combining label smoothing with PSL loss in a weighted manner. We refer to this variant as \textbf{CONNA+LS+PSL}.


\vpara{Implementation Details.}
We adopt Qwen3‑8B~\cite{yang2025qwen3} as the backbone and apply LoRA~\cite{edward2022lora} for parameter‑efficient fine‑tuning, using a rank of 16 and an alpha value of 32. We use the AdamW optimizer and 
a batch size of 1 with 8× gradient accumulation.

For the LLM-based Chain-of-Refinement process, we employ GPT‑5 to generate labels and set the profile‑cleaning batch size to 20. To improve robustness, we incorporate a retry mechanism with up to three attempts to recover from unparseable outputs resulting from occasional model instability. 
In addition, $\rho$ is set to $0.5$ to partition consistent set $\mathcal{D}_{\text{\mcs}}$ and inconsistent set $\mathcal{D}_{\text{\mics}}$.

Following the supervised fine‑tuning setup in~\cite{pang2025guard}, we construct the input context by concatenating the title, author, and affiliation information for each paper. For every target paper, we append a special token, \texttt{<label\_token>} to signal that a label should be predicted. The normal versus abnormal score is derived from the softmax‑normalized log probabilities of the ``Yes'' and ``No'' tokens. In addition, we adopt a multi‑turn instruction‑tuning strategy that enables the model to jointly predict labels for ten papers from the same author within a single inference round. We run each method five times and calculated the statistical significance of the differences between CrossND and baselines.



\begin{table}[t!]
\centering
\caption{Results of paper-author checks ($\%$). \textnormal{* denotes statistically significant improvement over the best baseline ($p<0.05$).}}
\label{tab:main_results}
\begin{tabular}{lcccc}
\toprule
\textbf{Method} & \multicolumn{2}{c}{\textbf{WhoIsWho}} & \multicolumn{2}{c}{\textbf{KDD Cup}} \\
\cmidrule(lr){2-3} \cmidrule(lr){4-5}
 & AUC & MAP & AUC & MAP \\
\midrule
GBDT        & 61.27 & 45.27 & 68.12 & 61.08 \\
RND-all & 46.37 & 45.76 & 20.92 & 28.73 \\
AuthorSim   & 71.14 & 51.74 & 50.56 & 20.72 \\
LOF         & 32.96 & 24.21 & 29.22 & 27.85 \\
DCI         & 53.27 & 34.46 & 58.61 & 40.22 \\
KEnS        & 45.97 & 29.43 & 53.84 & 40.01 \\
TK          & 57.60 & 37.02 & 56.55 & 30.27 \\
CONNA       & 59.61 & 44.27 & 66.53 & 61.12 \\
\midrule
LS          & 73.93 & 59.41 & 76.89 & 67.83 \\
MAE         & 74.47 & 58.91 & 76.17 & 67.89 \\
Co-teaching & 73.21 & 56.86 & 78.70 & 70.35 \\
CORES       & 74.54 & 58.96 & 76.63 & 69.14 \\
\midrule
DeepSeek-R1 &63.24  & 38.75 & 75.26 & 61.60 \\
GPT-5        &61.65  & 43.85 & 76.53 & 65.46 \\
Claude-Haiku-4.5 &67.91  & 44.04 & 78.55 & 65.95 \\
Gemini-3-Pro &76.11  & 54.12 & 77.39 & 67.32 \\      
GuARD       &77.14  & 55.02 & \underline{85.27} & 73.71 \\
\midrule
\textbf{CONNA+LS+PSL} & \underline{77.66} & \underline{61.31} & 83.74 & \underline{75.23} \\
\textbf{\model} & \textbf{82.19}* & \textbf{61.79}* & \textbf{87.34}* & \textbf{78.13}* \\
\bottomrule
\end{tabular}
\end{table}

\subsection{Main Results}
We compare the performance of our method against a variety of baselines
spanning traditional machine learning models, neural matching architectures, self-correction methods, and LLM-based approaches, as summarized in Table \ref{tab:main_results}.


Traditional machine learning methods (e.g., GBDT, LOF, and DCI) perform substantially poorly on both datasets, primarily because they depend on handcrafted features and lack the capacity to capture complex semantic or relational patterns in noisy environments. Both GBDT and RND-all underperform relative to advanced LLMs (e.g. Claude), highlighting the effectiveness of the semantic correlations reflected in LLMs. 

AuthorSim performs poorly on the KDD Cup dataset, suggesting that naively treating paper-author pairs with low cross-source author-similarity scores as incorrect assignments is not suitable for real-world settings. Similarly, anomaly detection approaches such as LOF and DCI yield unsatisfactory results, indicating that directly applying feature- or graph-structure-based anomaly detectors is ineffective for this task.
KEnS also fails to achieve competitive performance, implying that modeling paper–author matching purely through knowledge-graph embeddings is overly simplistic. 


Neural matching and noise‑robust self-correction learning methods, including LS, MAE, Co‑teaching, and CORES achieve markedly stronger performance via 
specialized noise‑handling strategies. 
However, 
their effectiveness is hindered by imperfect data and the low expressiveness of shallow networks. CONNA+LS+PSL outperforms LS, indicating that the PSL loss remains effective even in shallow neural network architectures.



We further evaluate several LLM-based approaches, including DeepSeek-R1, GPT‑5, Gemini-3-pro, and Claude-Haiku-4.5. Overall, these LLM-based methods exhibit limited effectiveness, suggesting that even highly capable LLM APIs struggle with batch semantic anomaly detection in complex relational environments. Compared with fine-tuned models, API-based LLMs still lack a robust inductive bias for accurate name disambiguation and error correction, resulting in limited predictive reliability.


In contrast, for our framework, in the chain-of-refinement stage, we mitigate noise in author-paper matching labels through profile cleaning, assignment prediction, and batch score refinement. During fine-tuning, we further reduce label noise by performing cross-source probabilistic soft logic inference, which leverages external data to estimate the matching probability of internal author-paper pairs. Together, these steps substantially enhance the reliability of the training labels and prediction scores.


\begin{table}[t]
\centering
\caption{Ablation studies. \textnormal{w/o out: without outer source. PC: Profile Cleaning.}}
\label{tab:ablation}
\begin{tabular}{p{1.2cm} p{2.1cm}cccc}
\toprule
\textbf{Setting} & \textbf{Model Variant} & \multicolumn{2}{c}{\textbf{WhoIsWho}} & \multicolumn{2}{c}{\textbf{KDD Cup}} \\
\cmidrule(lr){3-4} \cmidrule(lr){5-6}
 &  & AUC & MAP & AUC & MAP \\
\midrule
 All & \model &  \textbf{82.19}  &  \textbf{61.79} & \underline{87.34}   & \textbf{78.13} \\
\hline
\multirow{4}{*}{Stage 1+2}
& w/o TTS & \underline{81.11} & \underline{61.39} & \textbf{87.52} & 75.07 \\
 & w/o out & 79.57 & 56.34 & 86.41 & 72.26 \\
 & w/o PC  & 80.23 & 60.08 & 84.21 & 71.60 \\
 & w/o PSL & 80.93 & 58.29 & 86.74 & \underline{75.36} \\
\hline

\multirow{3}{*}{Stage 1}
 & w/o SFT          & 67.24 & 47.03 & 79.63 & 69.81 \\
 & \quad - Score Refine. & 65.74 & 46.36 & 78.19 & 67.79 \\
 & \quad - PC             & 61.65 & 43.85 & 76.53 & 65.46 \\
\bottomrule
\end{tabular}
\end{table}

\subsection{Ablation Studies}
We run ablations to measure the effect of each module.
Table~\ref{tab:ablation} reports the resulting performance on the two datasets.


We test the impact of four modules: TTS, outer source, profile cleaning (PC), and PSL-based cross‑correction. Removing TTS degrades performance on both datasets, confirming its robustness. Removing the outer source (w/o out) causes the largest degradation, with MAP drops of 5.05 on WhoIsWho and 2.81 on KDD Cup, showing that cross‑source signals are crucial for spotting incorrect assignments beyond what the internal source alone can reveal. Removing PC yields moderate declines, indicating that filtering anomalous papers improves downstream predictions. The effect of removing PSL varies across metrics: on WhoIsWho, MAP decreases by 3.10 and AUC slightly drops; on KDD Cup, AUC decreases but MAP improves marginally. These results suggest that PSL’s structured constraints can help ranking‑based metrics, while their impact on binary classification depends on dataset characteristics.


To evaluate the cumulative effect of our chain‑of‑refinement pipeline, we progressively remove components starting from supervised fine‑tuning (SFT). Removing SFT (w/o SFT) causes a major drop in MAP (–11.26 on WhoIsWho, –5.55 on KDD Cup). Eliminating the batch score refinement step yields further declines,
and removing profile cleaning produces the worst results. 
The gradual decline in performance demonstrates that every stage contributes substantially to the final results,
highlighting the importance of the full multi‑stage design in chain-of-refinement pipeline.

\begin{table}[t]
\centering
\caption{Effect of internal and external profiles.}
\label{tab:external_info}
\begin{tabular}{lcccc}
\toprule
\textbf{Stage-1 Setting} & \multicolumn{2}{c}{\textbf{WhoIsWho}} & \multicolumn{2}{c}{\textbf{KDD Cup}} \\
\cmidrule(lr){2-3} \cmidrule(lr){4-5}
 & \textbf{AUC} & \textbf{MAP} & \textbf{AUC} & \textbf{MAP} \\
\midrule
Cross-source & \textbf{65.74} & \textbf{46.36} & \textbf{78.19} & \textbf{67.79} \\
Internal-only & 64.21 & 41.30 & 77.76 & 66.20 \\
External-only & 48.58 & 25.49 & 77.06 & 65.29 \\
\bottomrule
\end{tabular}
\end{table}

To better understand the role of external-source data in incorrect assignment detection, we conduct an additional controlled comparison. 
Under the same Stage-1 protocol (Profile Cleaning + Assignment Prediction), we vary the profiles provided to the LLM. 
As shown in Table~\ref{tab:external_info}, the cross-source setting consistently outperforms both internal-only and external-only variants, indicating that the gains do not come from directly trusting the external source. 
The weak performance of the external-only variant, especially on WhoIsWho, further suggests that external assignments can also be noisy or only partially aligned with the target internal author. 
These results show that \model benefits from reconciling source agreement and disagreement rather than treating either source as ground truth.
We further evaluate different multi-turn assignment strategy in TTS, as shown in Appendix~\ref{app:tts_strategies}.


\hide{
\begin{table}[t]
\centering
\caption{Ablation studies. \textnormal{w/o out: without outer source. PC: Profile Cleaning.}}
\label{tab:ablation}
\begin{tabular}{lcccc}
\toprule
\textbf{Model Variant} & \multicolumn{2}{c}{\textbf{WhoIsWho}} & \multicolumn{2}{c}{\textbf{KDD Cup}} \\
\cmidrule(lr){2-3} \cmidrule(lr){4-5}
                               & AUC & MAP & AUC & MAP \\
\midrule
\model                     & \textbf{81.11} & \textbf{61.39} & \textbf{87.52} & 75.07 \\
\hline
w/o out                        & 79.57 & 56.34 & 86.41 & 72.26 \\
w/o PC                         &  80.23  &  60.08  & 84.21 & 71.60 \\
w/o PSL                        & 80.93 & 58.29 & 86.74 & \textbf{75.36} \\
\hline
w/o SFT                        &  67.24  &  47.03   &79.63&   69.81  \\
\quad  - Score Refine.            &  65.74  &  46.36   &78.19&   67.79\\
\quad  - PC                    &  61.65  &  43.85   &76.53&   65.46  \\
\bottomrule
\end{tabular}
\end{table}
}

\subsection{Hyperparameter Sensitivity}

We further examine the sensitivity of the hyperparameter $\lambda$, which controls the contribution of the PSL-based logical loss relative to the cross-entropy loss, and the margin parameter $\varphi$, which controls the separation boundary between satisfied and violative triples.

\begin{figure}[t]
\centering
\begin{subfigure}{0.48\columnwidth}
    \centering
    \includegraphics[width=\linewidth]{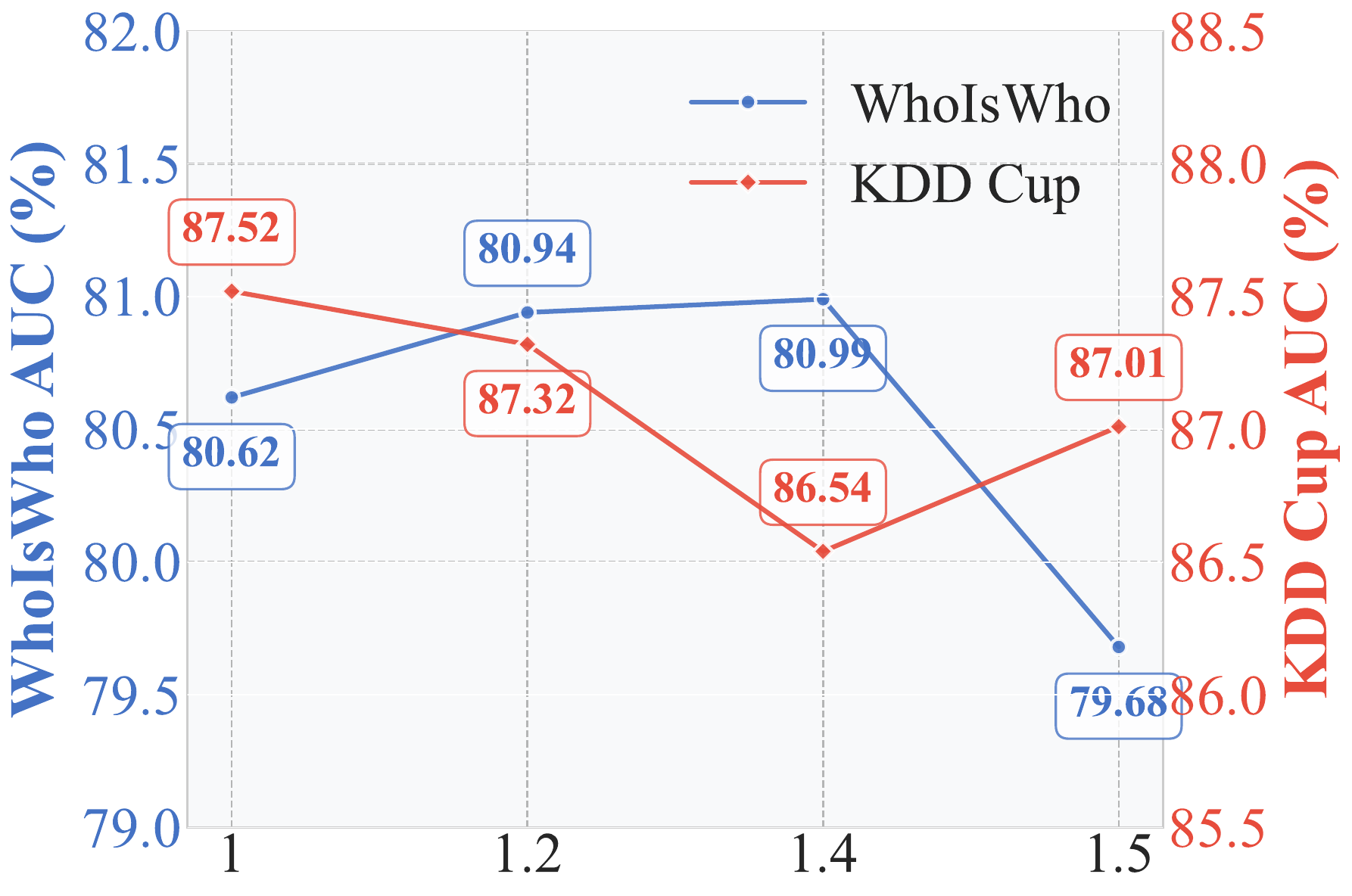}
    \caption{$\varphi$ with fixed $\lambda=0.5$}
    \label{fig:hyper_phi}
\end{subfigure}
\hfill
\begin{subfigure}{0.48\columnwidth}
    \centering
    \includegraphics[width=\linewidth]{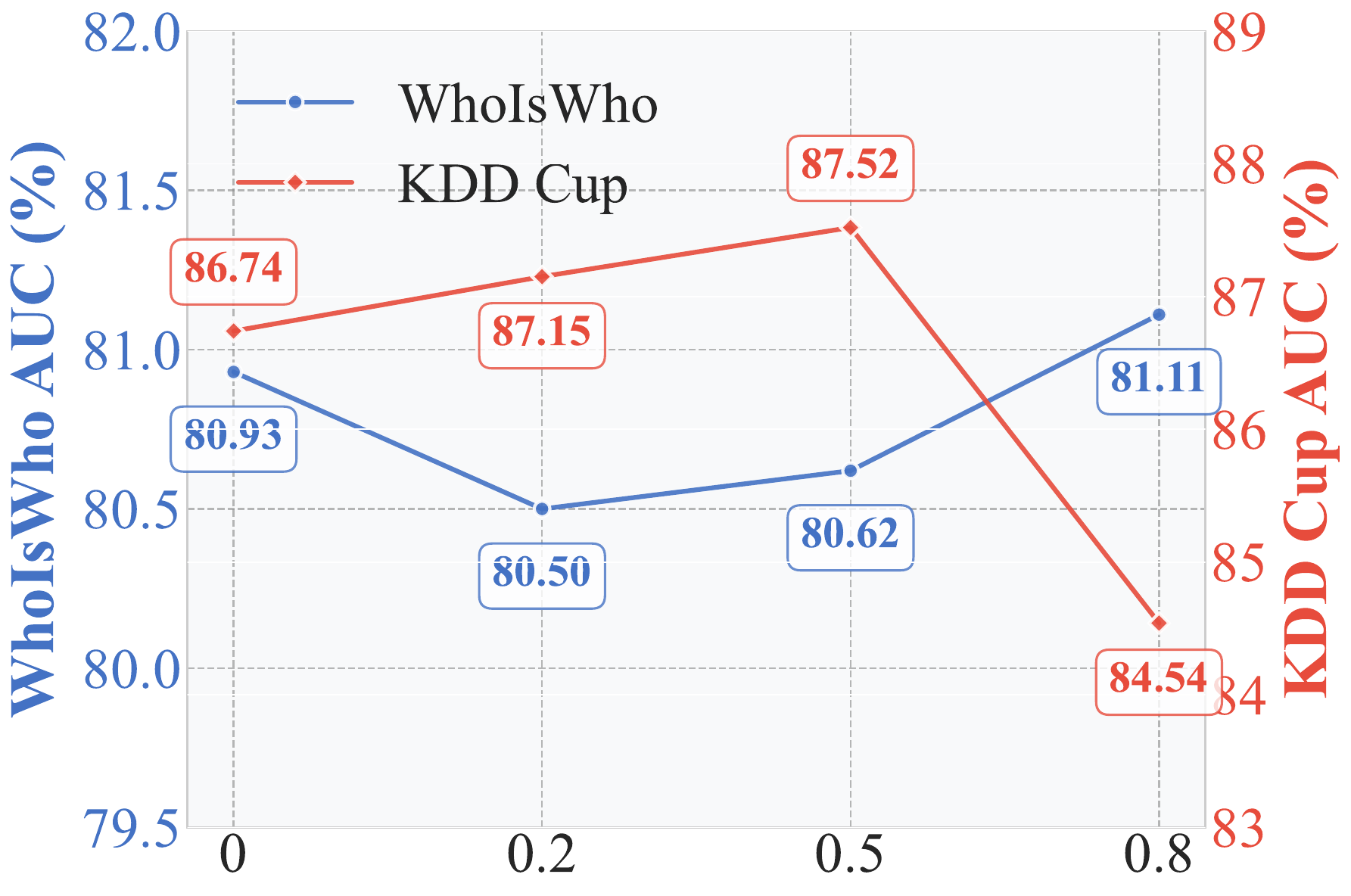}
    \caption{$\lambda$ with fixed $\varphi=1.0$}
    \label{fig:hyper_lambda}
\end{subfigure}
\caption{Effect of the margin $\varphi$ and the weight $\lambda$ of PSL loss.}
\label{fig:hyper_sensity}
\end{figure}

Figure~\ref{fig:hyper_phi} presents the results with a fixed loss weight $\lambda=0.5$. On WhoIsWho, the best performance is achieved at $\varphi=1.4$ with 80.99 AUC. On the KDD Cup dataset, $\varphi=1$ yields the best AUC of 87.52. These results indicate that moderate margin values around 1.0 work well across both datasets.

Figure~\ref{fig:hyper_lambda} reports model performance for varying $\lambda$ with the margin fixed at $\varphi=1$. On WhoIsWho, the optimal performance is achieved at $\lambda=0.8$ with 81.11 AUC. In contrast, the KDD Cup dataset exhibits the best AUC of 87.52 when $\lambda=0.5$. These results indicate that the optimal trade‑off between logical constraints and supervised signals differs across datasets.


\begin{figure}[H]
    \centering
    \includegraphics[width=1.05\linewidth]{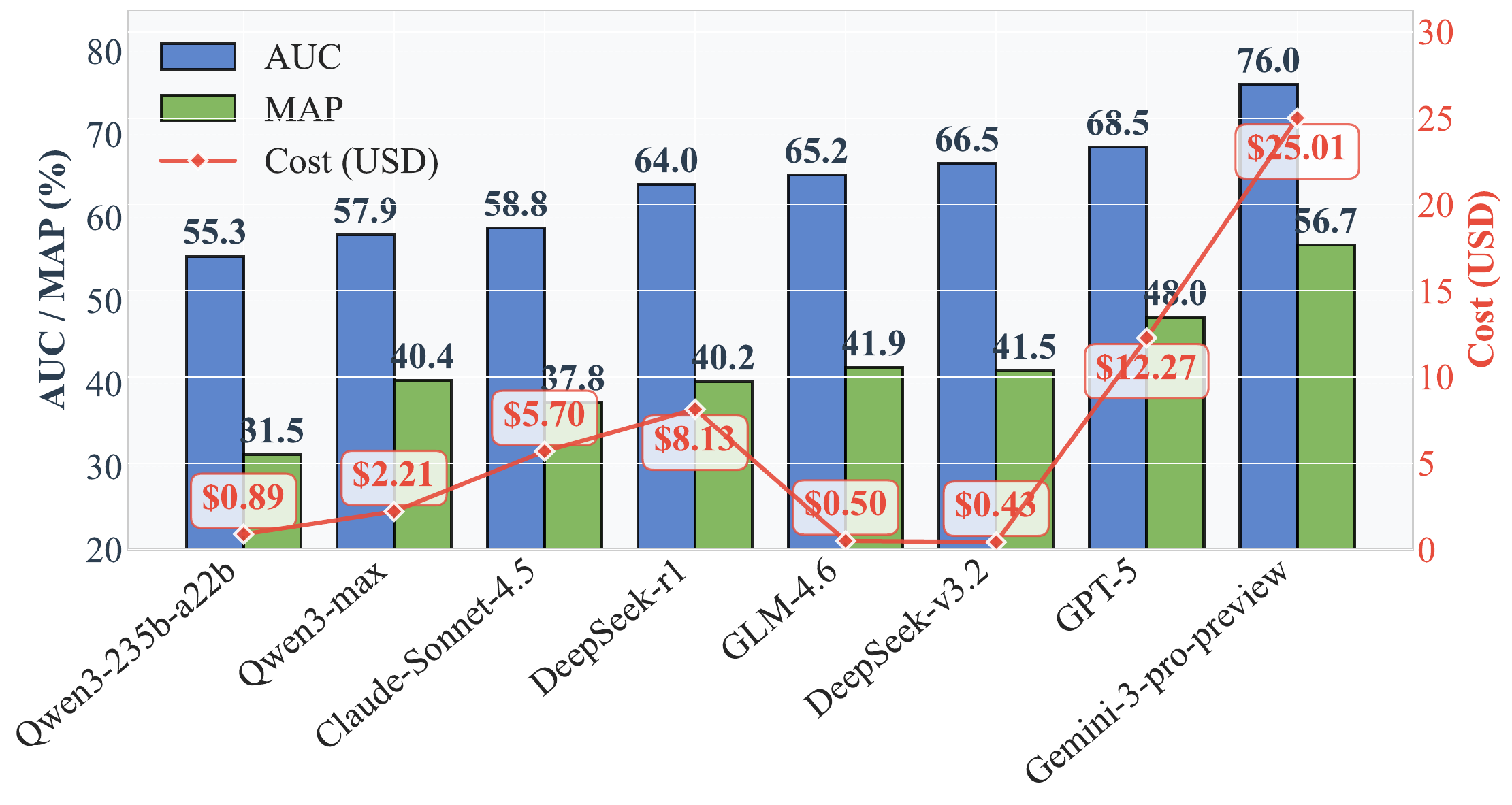}
    \caption{Comparison of different LLM APIs.}
    \label{fig:llms_cmp}
\end{figure}
 
\subsection{Performance of Different LLM APIs}

We benchmark the cross‑source disambiguation ability of leading LLMs on WhoIsWho. We randomly sample 50 authors ($\approx$ 5,000 papers) from the test set and construct batch‑query prompts for each model. We evaluate DeepSeek‑R1, DeepSeek‑V3.2~\cite{liu2025deepseek}, GLM‑4.6~\cite{zeng2025glm}, Qwen3‑Max~\cite{yang2025qwen3}, Qwen3‑235B‑A22B, Claude‑Sonnet‑4.5, GPT‑5, and Gemini‑3‑Pro, and record both accuracy and API cost. The results are shown in Figure~\ref{fig:llms_cmp}.


\begin{figure*}[t]
	\centering
	\includegraphics[width=0.95\textwidth]{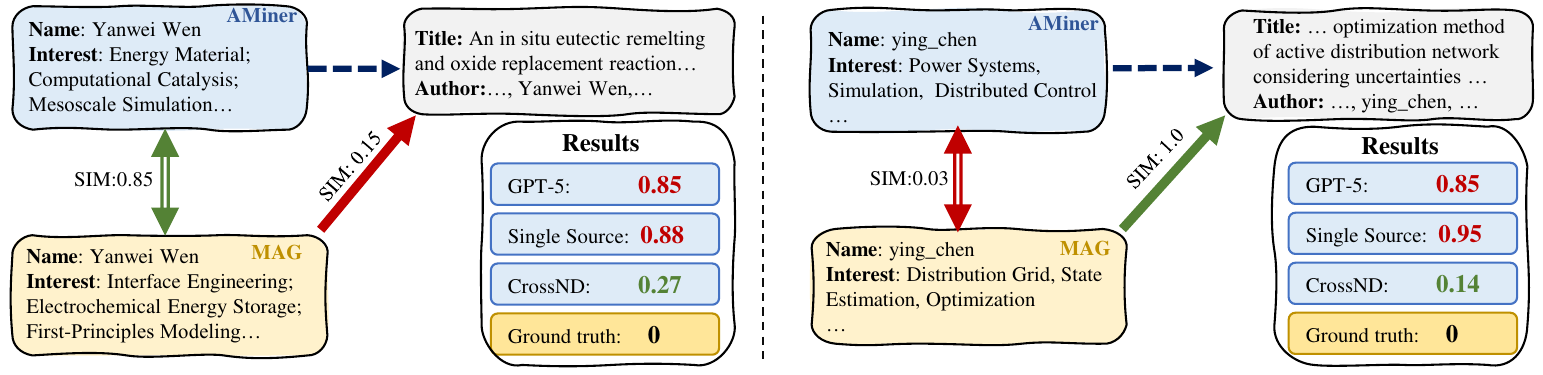}	
	\caption{Representative case studies. \model effectively leverages cross-source evidence to 
    detect incorrect assignments.}
	\label{fig:case_study}
\end{figure*}

\begin{table}[t]
\centering
\caption{Post-cutoff cross check on AMiner-Google Scholar papers published in 2024--2026.}
\label{tab:post_cutoff}
\begin{tabular}{lcc}
\toprule
\textbf{Method} & \textbf{AUC} & \textbf{MAP} \\
\midrule
GPT-5 & 63.42 & 49.21 \\
+ Profile Cleaning & 65.77 & 52.62 \\
+ Score Refinement & 65.16 & 54.50 \\
\model & 68.33 & 55.51 \\
\model+TTS & \textbf{70.79} & \textbf{58.35} \\
\bottomrule
\end{tabular}
\end{table}

The results highlight a clear performance–cost trade-off. \textbf{Gemini-3-Pro-Preview} delivers the highest accuracy (AUC 76.04) but at the highest cost (\$25.01). In contrast, \textbf{DeepSeek-V3.2} and \textbf{GLM-4.6} offer excellent cost-efficiency, achieving competitive AUC (66.53 and 65.18) for only \$0.43 and \$0.50. The reasoning-oriented \textbf{DeepSeek-R1} incurs substantial cost (\$8.13) due to its long outputs, even exceeding \textbf{Claude-Sonnet-4.5} (\$5.70). Overall, while premium models like GPT-5 and Gemini maximize accuracy, \textbf{DeepSeek-V3.2} emerges as the most practical option for large-scale, budget-sensitive disambiguation tasks.

\subsection{Evaluation on Post-cutoff Data}
To examine possible data leakage, we build a new AMiner dataset with papers published in 2024--2026 and use Google Scholar as the external source. It contains \num{1188} paper-author pairs, including \num{1000} positives and 188 negatives. This setting is challenging because Google Scholar metadata are less standardized and often lack affiliations or organizations. 
Table~\ref{tab:post_cutoff} shows the results. \model improves GPT-5 by 4.91 AUC and 6.30 MAP, and \model+TTS further increases the gains to 7.37 AUC and 9.14 MAP. These results suggest that the improvements are not tied to older benchmarks.

\subsection{Deployment and Practical Efficiency}

\hide{
\begin{table}[t]
\centering
\caption{End-to-end API cost of the chain-of-refinement pipeline.}
\label{tab:api_cost_small}
\resizebox{\linewidth}{!}{
\begin{tabular}{lcccc}
\toprule
\textbf{Stage} & \multicolumn{2}{c}{\textbf{WhoIsWho}} & \multicolumn{2}{c}{\textbf{KDD Cup}} \\
\cmidrule(lr){2-3} \cmidrule(lr){4-5}
 & \textbf{Tokens (M, in/out)} & \textbf{Cost (\$)} & \textbf{Tokens (M, in/out)} & \textbf{Cost (\$)} \\
\midrule
Profile cleaning & 56.65 / 1.89 & 89.75 & 14.31 / 0.59 & 23.88 \\
Assignment prediction & 48.92 / 2.16 & 82.79 & 9.55 / 0.29 & 14.85 \\
Batch score refinement & 29.46 / 1.63 & 53.14 & 1.48 / 0.21 & 3.94 \\
\midrule
Total & 135.03 / 5.68 & 225.68 & 25.35 / 1.10 & 42.67 \\
Average per paper & 0.0011 / 4e-5 & 0.0019 & 8e-4 / 3e-5 & 0.0013 \\
\bottomrule
\end{tabular}
}
\end{table}
}

\begin{table}[t]
\centering
\caption{End-to-end API cost of the chain-of-refinement pipeline.}
\label{tab:api_cost}
\setlength{\tabcolsep}{5pt}
\renewcommand{\arraystretch}{1.15}
\begin{tabular}{@{}lcccc@{}}
\toprule
\textbf{Stage} 
& \multicolumn{2}{c}{\textbf{WhoIsWho}} 
& \multicolumn{2}{c}{\textbf{KDD Cup}} \\
\cmidrule(lr){2-3} \cmidrule(lr){4-5}
& \makecell[c]{\textbf{Tokens}\\\textbf{(M, in/out)}} 
& \makecell[c]{\textbf{Cost}\\\textbf{(\$)}} 
& \makecell[c]{\textbf{Tokens}\\\textbf{(M, in/out)}} 
& \makecell[c]{\textbf{Cost}\\\textbf{(\$)}} \\
\midrule
Profile cleaning 
& 56.65 / 1.89 & 89.75 
& 14.31 / 0.59 & 23.88 \\
Assign. prediction 
& 48.92 / 2.16 & 82.79 
& 9.55 / 0.29 & 14.85 \\
Batch score refine. 
& 29.46 / 1.63 & 53.14 
& 1.48 / 0.21 & 3.94 \\
\midrule
Total 
& 135.03 / 5.68 & 225.68 
& 25.35 / 1.10 & 42.67 \\
Avg. per paper 
& 0.0011 / 4e-5 & 0.0019 
& 8e-4 / 3e-5 & 0.0013 \\
\bottomrule
\end{tabular}
\end{table}

Since \model is designed for large-scale correction and reviewer assistance rather than strict real-time serving, we evaluate its practical efficiency in terms of API cost and throughput. 
Table~\ref{tab:api_cost} reports the end-to-end API cost of the chain-of-refinement pipeline.
Overall, the average cost per paper is only \$0.0019 on WhoIsWho and \$0.0013 on KDD Cup. 
With respect to throughput, GPT-5 processes about 3.6 papers per second with single-threaded API calls for each chain-of-refinement step, while the SFT model reaches about 26 papers per second on a single H100-80G GPU. 
In practice, the more expensive chain-of-refinement stage can be run offline, while the deployed system mainly uses the SFT model to support online reviewer verification. 
Thus, the practical bottleneck is better characterized by throughput and scalability than by latency.

\subsection{Case Study}

In Figure~\ref{fig:case_study}, we present two representative cases demonstrating the effectiveness of our cross-correction framework. In the left example, the internal and external profiles are highly consistent ($\text{SIM}=0.85$). Both GPT-5 and the single-source baseline assign near-one scores and fail to identify the mismatch, whereas \model leverages the consistent cross-source signals and assigns a confident score of 0.27.  
In contrast, the right example exhibits low profile similarity, indicating conflicting evidence and an incorrect assignment in at least one source. GPT-5 and the single-source baseline are misled by the individual profiles, while our PSL-based framework propagates beliefs across sources to detect contradictions, yielding more reliable disambiguation decisions.

\subsection{Prototype System}
\begin{figure}[h]
    \centering
    \includegraphics[width=0.99\linewidth]{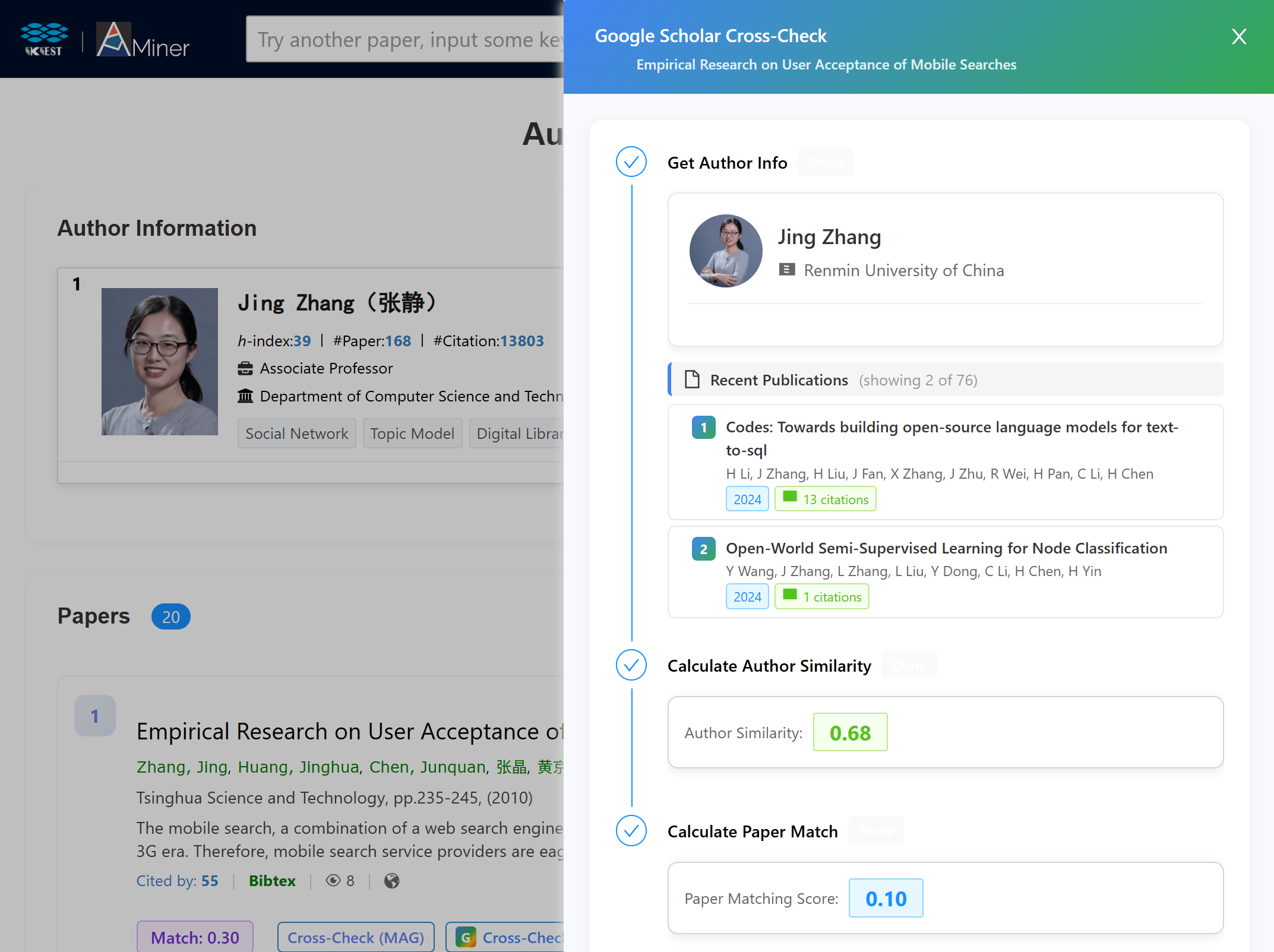}
    \caption{UI of the deployed system for comparing paper-author assignments across AMiner, MAG and Google Scholar.}
    \label{fig:na-demo}
\end{figure}

We have developed a prototype system\footnote{\url{https://na-demo.aminer.cn/crosscheckauthor}, launched by the end of November 2025.}
 for cross-correction across AMiner, MAG, and Google Scholar,
aimed at assisting users with error correction. 
A system screenshot is shown in Figure \ref{fig:na-demo}.
The target author is shown on the left. Below are the author’s papers in AMiner. Users can click \textit{Cross-check (MAG)} or \textit{Cross-check (Google Scholar)} to inspect the corresponding matched external author profiles. Based on an LLM fine-tuned on the WhoIsWho dataset, we present internal paper-author similarity, cross-source author similarity, and paper-to-external-author similarity to assist reviewers. Users can make more credible decisions by comparing the disambiguation results of the two sources. Experienced annotators have found \num{1061} assignment errors among \num{2154} paper-author pairs. Compared with the previous checking system, the average number of incorrect relationships detected per author increased by 47\%. 

\section{Related Work}

\subsection{Name Disambiguation}

From-scratch name disambiguation has been widely studied~\cite{fan2011graph,zhang2017name,cheng2024bond}. Early work relies on hand-crafted features to measure paper similarity~\cite{huang2006efficient,louppe2016ethnicity}, while later approaches learn paper embeddings from similarity graphs~\cite{zhang2018name,wang2020author}. Real-time disambiguation has also been explored; for example, CONNA~\cite{chen2022conna} models cross-field interactions and includes a decision module to handle non-existent authors. More recently, fine-tuned LLMs have been used for incorrect assignment detection~\cite{zhang2024enhanced,pang2025guard}, achieving strong performance using only internal data. However, these methods ignore the potential benefits of external data and their disambiguation results. Human-in-the-loop approaches~\cite{wang2011adana,qian2020partner} have also been proposed, but they are costly and require domain expertise. Overall, existing methods are developed under different assumptions and are not directly applicable to our setting.



\subsection{Learning with Noisy Labels}

Deep neural networks (DNNs) are vulnerable to label noise~\cite{arpit2017closer,yuan2024hide}. Prior work mitigates this issue via robust loss functions~\cite{zhang2018generalized}, 
regularization~\cite{muller2019does},
and sample selection~\cite{malach2017decoupling}, typically following a self‑correction paradigm. Recent LLM‑based methods, such as LAFT~\cite{wang2023noise}, leverage LLMs to identify noisy samples and provide auxiliary supervision, while SiDyP~\cite{ye2025calibrating} calibrates LLM‑generated labels through simplex diffusion in embedding spaces. In contrast, we study a complementary setting: \emph{cross‑correction} via external knowledge. Our approach is orthogonal to self‑correction and can be seamlessly combined with existing methods.

\section{Conclusion}

This paper studies name disambiguation from a new perspective of cross‑source reasoning–based correction. We propose \model, a unified framework that first applies a chain‑of‑refinement pipeline to clean author profiles and integrate external signals, producing more reliable author–paper matching probabilities under noisy labels. Building on these refined signals, we introduce an SFT framework combined with a probabilistic soft logic (PSL) cross‑correction module to identify incorrect author–paper assignments and generate more robust training supervision. Extensive experiments demonstrate the effectiveness of our cross‑correction approach, offering new insights for advancing author name disambiguation. 






\begin{acks}
This work has been supported by the National Natural Science Foundation of China (62406164, 62372483, 62425601)  and New Cornerstone Science Foundation through the XPLORER PRIZE.
We also acknowledge the support from Public Computing Cloud, Renmin University of China.
It was also partially supported by the
EU Project SMARTY (GA 101140087).


\end{acks}

\bibliographystyle{ACM-Reference-Format}
\bibliography{ref}

\appendix

\section{Hallucination analysis of \model.}

Since Stage~1 uses LLM-generated explanations to refine soft labels, we further analyze whether these explanations introduce hallucinated or extra-contextual information that could bias the soft-label refinement process. In this setting, the key concern is that the LLM should not rely on its internal knowledge about scholars, authors, or institutional backgrounds, but should instead make decisions based only on the provided input context. For each explanation, we provide a GPT-5 judge with the original input and ask it to score three dimensions: context support, absence of external assumptions, and relevance focus. Each dimension is rated on a 1--5 scale, where a higher score indicates lower hallucination risk and stronger grounding in the input context.

\begin{table}[H]
\centering
\small
\caption{Hallucination analysis of LLM-generated explanations.}
\label{tab:hallucination}
\renewcommand{\arraystretch}{1.15}
\begin{tabular}{p{0.22\linewidth}p{0.48\linewidth}c}
\toprule
\textbf{Dimension} & \textbf{1--5 Scale Definition} & \textbf{Mean Score} \\
\midrule
\makecell[l]{Context\\Support}
& \makecell[l]{1 = unsupported/contradicted\\3 = partially supported\\5 = fully supported by the input}
& 4.758 \\
\midrule
\makecell[l]{Absence of\\External\\Assumptions}
& \makecell[l]{1 = relies on outside knowledge\\3 = some unsupported inference\\5 = no extra-contextual assumptions}
& 4.808 \\
\midrule
\makecell[l]{Relevance\\Focus}
& \makecell[l]{1 = largely irrelevant\\3 = partly relevant\\5 = focused on contextual evidence}
& 4.969 \\
\bottomrule
\end{tabular}
\end{table}


Table~\ref{tab:hallucination} shows that the generated explanations exhibit low hallucination risk across \num{1502} explanations. In particular, the high score for absence of external assumptions suggests that the model rarely relies on information beyond the provided input. We further inspect low-scoring cases and find that the main issue is not factual fabrication or external-knowledge leakage. Instead, most errors come from over-confident wording under limited evidence, such as describing two profiles as ``fully consistent'' when the evidence only partially supports the conclusion. Thus, the main risk lies in confidence calibration rather than typical hallucination or reliance on the LLM's internal scholar knowledge.

\section{Prompts Used in \model}
\label{app:appendix_prompt}

\vpara{Prompts of the Chain-of-Refinement model.} 
The system and user prompts of Profile Cleaning are:
\begin{lstlisting}[style=prompt]
System prompt: 
You are performing scholar-paper attribution anomaly detection. Each scholar has an internal paper set that may contain incorrectly assigned papers, and an external paper set used as supporting evidence. Your task is to determine whether each queried paper truly belongs to the target scholar by considering co-authors, research field, publication venue, and affiliation, as well as consistency between internal and external sources. Your output should be a JSON object in the form {"id":"label"}, where 0 means "does not belong" and 1 means "belongs".

User prompt:
Target scholar: {name}
Internal paper collection: {collection}
Papers to be judged: {multi_turn_input}

Assistant prompt:
{"id":"label"}
\end{lstlisting}
The system and user prompts of Assignment Prediction
are:
\begin{lstlisting}[style=prompt]
System prompt:  
You are performing scholar-paper anomaly attribution detection. For each target scholar, two paper collections are provided: an internal source and an external source.  The internal source is the target set for anomaly detection, which may contain wrongly assigned papers.  The external source is retrieved from external databases and is used as supporting evidence, but it may also contain some misassigned papers.Your task is to determine whether each paper in the internal source truly belongs to the target scholar by considering research field, affiliation, co-authors, and publication venues, as well as its similarity with papers in both the internal and external sources. For each target paper, you should output a confidence score between 0 and 1 indicating the likelihood that the paper belongs to the scholar. Your final output must be a valid JSON object of the form: {"0": 0.2, "1": 0.95, "2": 0.4, ...}, where the key is the paper ID and the value is a float in [0,1]. A score closer to 0 means the paper is likely not written by the scholar, while a score closer to 1 means it is likely written by the scholar. You should output only the JSON object, with no additional text, and it must be parsable by json.load.

User prompt:  
Target scholar: {name}  
Internal source: {inner}  
External source: {outer}  

Assistant prompt:  
{"id": score}
\end{lstlisting}
The system and user prompts of Batch Score Refinement
are:
\begin{lstlisting}[style=prompt]
System prompt:  
You are performing scholar-paper anomaly attribution detection. You will receive multiple batches of papers. Each paper includes metadata such as title, authors, affiliations, and publication venue, as well as a prior anomaly score in [0,1]. This prior score comes from a previous attribution model: values closer to 1 indicate that the paper is more likely to belong to the target scholar. However, scores from different batches may have different ranges and calibration, and they may contain noise or errors. You should infer the scholar's overall profile (e.g., main research fields, frequent collaborators, and institutional affiliations) from all provided papers and their prior scores. Based on this holistic understanding, you must re-evaluate each paper and assign a new confidence score indicating how likely it is to truly belong to the scholar. Your final output must be a single valid JSON object of the form: {"0": 0.12, "1": 0.87, "2": 0.43, ...}, where each key is a paper ID and each value is a float in [0,1]. A value closer to 0 means the paper is likely anomalous (does not belong), while a value closer to 1 means it likely belongs to the scholar.  

User prompt:  
Target scholar: {name}  
Multi-batch papers with prior scores: {papers}

Assistant prompt:  
{"id": score}
\end{lstlisting}

\vpara{Prompts of the SFT model.} The system and user prompts of the SFT model are:

\begin{lstlisting}[style=prompt]
System prompt: 
You are performing scholar-paper attribution detection. Each scholar has two sources: an internal source and an external source, each containing a set of papers (with title, authors, and affiliation). Use the external source to support error detection in the internal source. Your task is to determine whether a given paper truly belongs to the target scholar. Papers that do not belong are considered anomalous."

User prompt:
Target scholar: {name}
Internal source: {inner}
External source: {outer}
Similarity between internal and external sources: {author_sim}"

"Is the following paper written by this scholar? {paper}"

Assistant prompt: "{label_token}"

...

"Is the following paper written by this scholar? {paper}"

Assistant prompt: "{label_token}"
\end{lstlisting}

\section{Test-Time Scaling Strategies} 
\label{app:tts_strategies}

\begin{table}[H]
\centering
\small
\caption{Ablation study of Test-Time Scaling strategies on the WhoIsWho dataset. Conf. denotes Confidence.}
\label{tab:tts_results}
\begin{tabular}{>{\centering\arraybackslash}m{1.6cm}|>{\centering\arraybackslash}m{1.2cm}|l|c|c}
\toprule
\textbf{Cross-batch} & \textbf{Signal} & \textbf{Within-batch} & \textbf{AUC} & \textbf{MAP} \\
\midrule
Random & -- & Random & 81.11 & 61.39 \\
\midrule
\multirow{4}{*}{Contiguous} & \multirow{2}{*}{Score} & Desc & 81.36 & 57.89 \\
& & Asc & 82.05 & 59.63 \\
\cline{2-5}
& \multirow{2}{*}{Conf.} & Desc & 81.18 & 59.86 \\
& & Asc & 81.95 & 59.37 \\
\midrule
\multirow{6}{*}{Snake} & \multirow{3}{*}{Score} & Desc & 81.95 & 59.85 \\
& & HighLow\_Alternate & 81.67 & 61.60 \\
& & ThreeWay\_Alternate & 81.38 & 60.22 \\
\cline{2-5}
& \multirow{3}{*}{Conf.} & Desc & 81.98  & 60.17 \\
& & HighLow\_Alternate & \textbf{82.19} & \textbf{61.79} \\
& & ThreeWay\_Alternate & 81.98 & 61.47 \\
\bottomrule
\end{tabular}
\end{table}

We compare different strategies for our Test-Time Scaling approach. As described in Section \ref{sec:tts}, our method consists of three components: confidence estimation, cross-batch balancing, and within-batch interleaving. Here we ablate design choices for each component.

For \textbf{confidence estimation}, we compare two signals: \textit{Score} uses the raw soft predictions $\bar{y}_i$; \textit{Conf.} uses confidence $c_i = |\bar{y}_i - 0.5|$.

For \textbf{cross-batch balancing}, we evaluate two assignment methods: \textit{Contiguous} directly partitions sorted papers sequentially into batches; \textit{Snake} uses the snake dealing method to distribute papers evenly across batches.

For \textbf{within-batch interleaving}, we test four ordering patterns: \textit{Desc} (descending by confidence), \textit{Asc} (ascending), \textit{Interleaving} (alternating high-low pattern), and \textit{ThreeWay} (interleaving high-mid-low three groups).

Table~\ref{tab:tts_results} shows the results. The Contiguous assignment strategy consistently underperforms the random baseline, with MAP scores ranging from 57.89 to 59.86 compared to the random baseline's 61.39. This substantial degradation demonstrates that assigning papers with similar scores to the same batch harms performance, likely because it reduces the diversity of anchors available for in-context learning within each batch.

In contrast, the Snake distribution strategies show much stronger performance. Within the Snake strategies, confidence-based signals consistently outperform score-based signals: the best confidence-based result (MAP: 61.79) exceeds the best score-based result (MAP: 61.60), and confidence-based strategies achieve more robust performance across different within-batch orderings (MAP range: 60.17-61.79 vs. 59.85-61.60 for score-based). This suggests that confidence captures more useful information for identifying effective anchor-target pairs.

Among within-batch ordering patterns, Interleaving consistently achieves the best MAP scores for both signals under Snake distribution. The optimal configuration (\textit{Snake + Conf. + Interleaving}) achieves AUC of 82.19 and MAP of 61.79, substantially outperforming the random baseline (AUC: 81.11, MAP: 61.39) and demonstrating that strategic test-time scaling can enhance model performance without additional training.

\end{document}